\documentclass[11pt]{article}
\usepackage[preprint]{acl}
\usepackage{times}
\usepackage[T1]{fontenc}
\usepackage[utf8]{inputenc}
\usepackage{microtype}
\usepackage{inconsolata}
\usepackage{graphicx}
\graphicspath{{Figures/}}
\usepackage{amsmath}
\usepackage{algorithm}
\usepackage{algpseudocode}
\usepackage{booktabs}
\usepackage{multirow}
\usepackage{colortbl}
\usepackage{tabularx}
\definecolor{searchweaverblue}{HTML}{EAF3FC}

\newcommand{\Ours}{SearchWeaver}

\title{Let Agents Design How They Search:\\
Automatic Interface Optimization for Agentic Search}

\author{
  \textbf{Sangjune An}$^{1*}$ \quad
  \textbf{Jaeyoon Jung}$^{1*}$ \quad
  \textbf{Yumin Choi}$^{2*}$ \\
  \textbf{Jinheon Baek}$^{2}$ \quad
  \textbf{Junmo Cho}$^{2}$ \quad
  \textbf{Sung Ju Hwang}$^{2}$ \quad
  \textbf{Hae Beom Lee}$^{1\dagger}$ \\[2pt]
  $^{1}$Korea University \quad $^{2}$KAIST \\
  \texttt{\{ics8088,tonyjung01,haebeomlee\}@korea.ac.kr}
  \quad
  \texttt{yuminchoi@kaist.ac.kr}
}

\begin{document}

\maketitle
\makeatletter
\ifacl@finalcopy
  \renewcommand{\thefootnote}{\fnsymbol{footnote}}
  \footnotetext[1]{Equal contribution; $^{\dagger}$Corresponding author.}
  \renewcommand{\thefootnote}{\arabic{footnote}}
\fi
\makeatother
\begin{abstract}
Agentic search depends on the interface through which a language-model agent accesses and explores a corpus.
Existing interface designs are fixed by hand and applied unchanged to every corpus, although effective designs may depend on the corpus.
We introduce \Ours{}, a framework for automatic search interface optimization guided by downstream search performance.
A proposer uses execution trajectories and task rewards to jointly revise corpus artifacts, executable tools, and usage instructions.
This process optimizes what information is prepared in advance, which operations tools perform, and which decisions remain with the search agent.
With GPT-5.4-nano, the optimized interfaces improve answer accuracy from 71.6\% to 78.4\% on BrowseComp-Plus and average nDCG@10 from 45.7 to 55.0 across four BRIGHT subsets over Direct Corpus Interaction.
These findings highlight the value of jointly optimizing corpus artifacts, executable tools, and usage instructions to develop effective search interfaces.
\end{abstract}

\section{Introduction}

Language-model agents search for evidence by issuing queries, inspecting retrieved information, and deciding what to explore next \citep{yao2023react,asai2024selfrag}.
These decisions depend on the search interface provided to the agent: how documents are organized, which operations are available, and what instructions guide their use.
The interface determines what evidence the agent can access at each step and how it can act on that evidence.
Designing this interface is therefore central to both the quality and the cost of agentic search.

Existing approaches provide different interfaces for accessing a corpus.
Retrieval-augmented generation supplies passages selected by a retriever \citep{lewis2020rag}, while Direct Corpus Interaction (DCI) lets agents search and read raw documents with general-purpose terminal tools \citep{li2026dci}.
More recent approaches combine retrieval with direct exploration: DR-DCI dynamically expands a local workspace with retrieved documents, and RISE constructs a bounded interaction space for the agent \citep{lu2026drdci,zhuang2026rise}.
Each approach makes design choices about which information to expose and which operations the agent can perform.
However, these approaches generally specify an interface design in advance and apply it unchanged across corpora, despite the fact that the most effective design depends on corpus size and structure, the types of queries, and the language model used for search.

This raises the question of how to design a search interface for a given corpus and language-model agent.
The goal is to determine what corpus information to prepare, which search operations to expose, and how to guide their use so as to improve the agent's downstream search performance.
An effective search interface must align the information and operations it provides with how the agent uses them.
For example, a title index can help locate documents about a candidate entity, but its usefulness depends on tools that expose those documents and instructions that guide the agent to inspect them and verify the query's clues.
These components therefore need to be designed together.

To tackle this, we propose \Ours{}, a framework that iteratively optimizes search interfaces using feedback from agent search trajectories and task rewards.
Specifically, a proposer analyzes failures and bottlenecks in these trajectories to jointly revise which corpus artifacts to prepare, how to structure executable tools, and how to guide the agent in using them.
Each revised interface is evaluated on search tasks, producing new trajectories and rewards that guide subsequent revisions.
In this way, optimization shapes the division of work between corpus preparation, tool execution, and agent reasoning.

Experiments with GPT-5.4-nano show that the optimized interfaces improve answer accuracy from 71.6\% to 78.4\% on BrowseComp-Plus~\citep{chen2025browsecompplus} and average nDCG@10 from 45.7 to 55.0 across four BRIGHT~\citep{su2025bright} subsets over plain DCI, while reducing mean tool execution time by 42\% on BRIGHT.
The discovered interfaces use distinct strategies across corpora, including restricting retrieval to a topic directory and combining agent judgments with retrieval rankings.
The BRIGHT interfaces also improve search performance and reduce tool execution time when reused by Gemini~3.5 Flash-Lite without re-optimization.
These findings support treating the search interface as a reusable corpus resource that can be optimized through agent execution.

\section{Method}

\subsection{Problem Definition}

Given a fixed corpus $C$ and a language-model agent $f_\theta$, a search interface comprises the corpus artifacts, executable tools, and usage instructions through which the agent searches for evidence and produces task outputs.
It determines which operations are available, what information they return, and how the agent is guided to use them.
The interface can thus change the agent's actions and outcomes even when the model and corpus remain the same.

Let $\phi\in\Phi$ denote an admissible search interface design, and let $f_\theta(q;C,\phi)$ denote the agent's output for query $q$ using that design.
Given a set of optimization queries $\mathcal{Q}$, we seek
\begin{equation}
  \phi^\star = \underset{\phi\in\Phi}{\arg\max}\;
  \frac{1}{|\mathcal{Q}|}\sum_{q\in\mathcal{Q}}
  R\!\left(f_\theta(q;C,\phi),q\right),
  \label{eq:objective}
\end{equation}
where $R$ measures task success (answer accuracy or nDCG@10 in our experiments) and the model parameters $\theta$ remain fixed.
A single interface is reused across queries, while the agent chooses its actions in response to each query and the evidence it observes.
The optimization therefore seeks a reusable way of organizing and guiding search across queries.

\subsection{\Ours{}: Automatic Search Interface Optimization}

\paragraph{Search Interface Design.}
Starting from DCI, in which an agent explores raw documents with terminal tools \citep{li2026dci}, \Ours{} seeks the information and operations that best extend this interface for each corpus.
\Ours{} represents each candidate interface as corpus artifacts, executable tools, and usage instructions; a candidate may also replace the agent's base system prompt and its context-management policy.
Artifacts organize corpus information before query execution for reuse across queries; tools expose operations over these artifacts and raw documents; instructions guide the agent in using those operations.
Tools can use the supplied BM25~\citep{robertson2009bm25} and dense~\citep{karpukhin2020dense} retrieval resources, restrict exploration to a subset of the corpus, or combine retrieval rankings with the agent's document judgments.
Returned documents retain their raw corpus paths for source inspection.
These components are designed jointly: a tool that uses agent judgments also requires instructions specifying when to inspect documents and how to pass those judgments to the tool.
The search space thus couples corpus preparation with query-time interaction, allowing what is prepared to change with how the agent accesses and uses it.

\paragraph{Trajectory-Guided Revision.}
To explore this design space, we use a coding-agent proposal--evaluation loop inspired by Meta-Harness~\citep{lee2026metaharness}.
Whereas prompt optimizers revise text~\citep{khattab2024dspy,agrawal2026gepa} and Meta-Harness revises a model's harness code, \Ours{} revises the interface the agent is given: corpus artifacts, executable tools, and usage instructions.
A proposer inspects the corpus and previous candidates, writes a new interface, and receives feedback from the search agent's execution with that interface.
This feedback includes per-query rewards and full execution trajectories containing commands, tool outputs, and final responses.
Rewards indicate which queries a candidate handles well or poorly; trajectories show how the agent uses the interface and where its search fails.
Gold document identifiers for optimization queries provide additional diagnostic feedback.
The proposer uses this evidence to revise artifacts, tools, and instructions, changing what information is exposed, which operations are implemented in tools, and which judgments are requested from the agent.

\paragraph{Evaluation and Selection.}
Each candidate's artifacts and tools are prepared over the corpus before the fixed search agent runs on $\mathcal{Q}$.
We record the resulting trajectories and per-query rewards and make them available to the proposer for subsequent revisions.
Every evaluated candidate contributes this feedback, regardless of whether it improves the current best score.
After a fixed number of iterations, we select the candidate with the highest mean reward on $\mathcal{Q}$ and evaluate it on test queries, which never enter proposer feedback.
We provide the algorithm for \Ours{} in Appendix~\ref{sec:optimization-algorithm}.

\begin{table}[!t]
\centering
\fontsize{8}{9.5}\selectfont
\setlength{\tabcolsep}{1.5pt}
\begin{tabularx}{\columnwidth}{@{}l*{6}{>{\centering\arraybackslash}X}@{}}
\toprule
\multirow{2}{*}[-2.5pt]{Method} & \multirow{2}{*}[-2.5pt]{BCP} & \multicolumn{5}{c}{BRIGHT} \\
\cmidrule(lr){3-7} 
& & Bio. & Earth & Econ. & Rob. & Avg. \\
\midrule
DCI & 71.6 & 61.6 & 51.5 & 31.3 & 38.2 & 45.7 \\
Agentic Search (Dense) & 62.6 & 58.2 & 51.4 & 34.6 & 35.7 & 45.0 \\
Agentic Search (BM25) & 47.5 & 48.8 & 41.6 & 30.2 & 35.4 & 39.0 \\
RISE & 65.0 & 34.9 & 29.4 & 17.7 & 21.7 & 25.9 \\
DR-DCI & 73.1 & 54.4 & 43.4 & 31.7 & 39.0 & 42.1 \\
\rowcolor{searchweaverblue}
\Ours{} (Ours) & \textbf{78.4} & \textbf{74.6} & \textbf{64.6} & \textbf{38.3} & \textbf{42.4} & \textbf{55.0} \\
\bottomrule
\end{tabularx}
\vspace{-4pt}
\caption{Main results with GPT-5.4-nano. Accuracy (\%) on BCP and nDCG@10 on BRIGHT.}
\label{tab:main}
\vspace{-10pt}
\end{table}

\section{Experiments}
\label{sec:experiments}

\subsection{Experimental Setup}
\label{sec:experimental-setup}

\paragraph{Benchmarks and Metrics.}
We evaluate \Ours{} on BrowseComp-Plus (BCP; \citealp{chen2025browsecompplus}) for question answering and BRIGHT~\citep{su2025bright} for reasoning-intensive retrieval.
On BCP, the agent searches a fixed web corpus to produce a short answer, and we measure answer accuracy judged by GPT-4.1~\citep{openai2025gpt41}.
We use 30 queries for interface optimization and 800 for testing.
On BRIGHT, the agent returns a ranked list of documents, which we evaluate using nDCG@10~\citep{jarvelin2002cumulated}.
We use four subsets (Biology, Earth Science, Economics, and Robotics) with 213 optimization and 210 test queries in total.

\begin{figure}[!t]
\centering
\includegraphics[width=0.95\columnwidth]{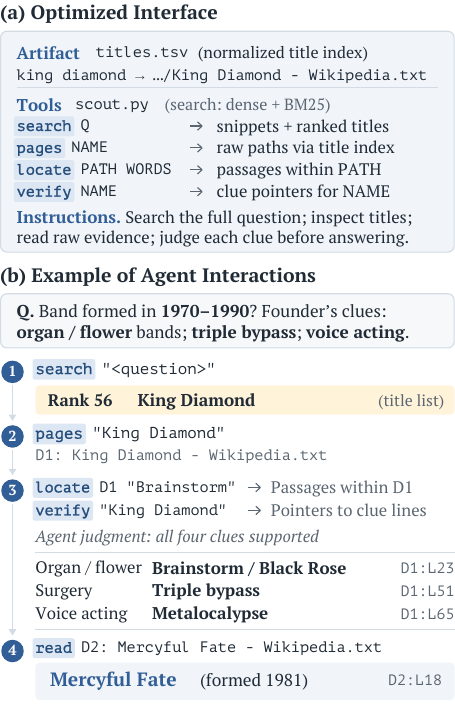}
\vspace{-6pt}
\caption{The optimized BrowseComp-Plus interface and selected agent interactions on a test query.}
\label{fig:bcp-qualitative}
\vspace{-10pt}
\end{figure}

\paragraph{Baselines.}
We compare against \textbf{DCI}~\citep{li2026dci}, which searches raw documents using a shell and file reader.
\textbf{Agentic Search (Dense)} and \textbf{Agentic Search (BM25)} iteratively reformulate queries for a search tool returning five document snippets per call~\citep{chen2025browsecompplus}.
The dense variant uses Qwen3-Embedding-8B~\citep{zhang2025qwen3embedding} on BCP and ReasonEmbed-Qwen3-4B~\citep{chen2026reasonembed} on BRIGHT.
\textbf{RISE}~\citep{zhuang2026rise} retrieves documents into a restricted workspace for shell-based exploration and augments documents with model-generated tables of contents and section anchors.
\textbf{DR-DCI}~\citep{lu2026drdci} lets the agent pull dense-retrieved documents into an initially empty workspace and explore them with shell commands.
\textbf{\Ours{}} uses execution feedback to automatically optimize a DCI agent's corpus artifacts, search tools, and usage instructions.

\paragraph{Implementation Details.}
Our search agent is implemented using Pi~\citep{zechner2026pi} with GPT-5.4-nano~\citep{openai2026gpt54nano}, following DCI~\citep{li2026dci}.
The proposer uses Claude Code~\citep{anthropic2025claudecode} with Claude Opus 5.5~\citep{anthropic2026opus55}.
We optimize an interface for each corpus, starting from the default DCI interface and evaluating one candidate interface per iteration for ten iterations.
We provide the proposer with BM25 and Qwen3-Embedding-4B retrieval on BCP, and BM25 and the released ReasonEmbed-Qwen3-8B rankings~\citep{chen2026reasonembed} on BRIGHT.
See Appendix~\ref{sec:experimental-details} for more implementation details.

\begin{figure}[!t]
\centering
\includegraphics[width=\columnwidth]{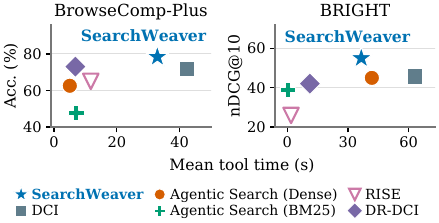}
\vspace{-6pt}
\caption{\Ours{} is Pareto-optimal among the evaluated methods on both benchmarks.}
\label{fig:performance-tool-time}
\vspace{-18pt}
\end{figure}

\subsection{Main Results and Analyses}
\label{sec:experimental-results}
\label{sec:analysis}

\paragraph{Search Performance.}
Table~\ref{tab:main} shows that \Ours{} outperforms all evaluated baselines on both BCP and BRIGHT, including each of the four BRIGHT subsets.
Compared with DCI, it improves BCP accuracy from 71.6\% to 78.4\% and average BRIGHT nDCG@10 from 45.7 to 55.0.
The relative performance of existing approaches varies across corpora: DR-DCI improves over DCI on BCP but falls below it on BRIGHT.
In contrast, \Ours{} improves over DCI on both benchmarks and all four BRIGHT subsets, supporting the value of optimizing interfaces for each corpus.

\paragraph{Discovered Interfaces.}
We examine the interfaces discovered by optimization for each corpus.
These interfaces differ in how they combine retrieval, source inspection, and agent judgments.
On BCP, as shown in Figure~\ref{fig:bcp-qualitative}, the generated title index maps candidate names to raw documents, and the tools support hybrid candidate retrieval and targeted evidence lookup.
The instructions guide the agent to use these operations to check each clue against source evidence.
The agent checks the clues about King Diamond against source lines and then confirms the formation year of Mercyful Fate from its own page.
We provide the full trajectory in Appendix~\ref{sec:bcp-trajectory}.
On BRIGHT, the interfaces target document ranking: Biology restricts retrieval to a topic directory, Earth Science introduces retrieval only when constructing the final ranking, and Economics and Robotics incorporate agent judgments through document ordering and relevance grades, respectively (Appendix~\ref{sec:bright-interfaces}).

\paragraph{Tool Execution Time.}
Figure~\ref{fig:performance-tool-time} shows that \Ours{} achieves these performance gains while using less tool execution time than DCI.
Mean tool execution time per query, which excludes model inference, decreases from 42.4\,s to 32.9\,s on BCP and from 63.2\,s to 36.6\,s on BRIGHT, reductions of 22\% and 42\%, respectively.
The reduction holds across all four BRIGHT subsets (Appendix~\ref{sec:bright-cost}).
These savings are consistent with the more targeted corpus access provided by the optimized interfaces.
For example, the Biology interface restricts retrieval to a topic directory, while the BCP interface uses a title index to locate candidate documents and passage retrieval to check specific clues.

\begin{table}[!t]
\centering
\fontsize{8}{9.5}\selectfont
\setlength{\tabcolsep}{1.5pt}
\begin{tabularx}{\columnwidth}{@{}l*{6}{>{\centering\arraybackslash}X}@{}}
\toprule
\multirow{2}{*}[-2.5pt]{Method} & \multirow{2}{*}[-2.5pt]{BCP} & \multicolumn{5}{c}{BRIGHT} \\
\cmidrule(lr){3-7} 
& & Bio. & Earth & Econ. & Rob. & Avg. \\
\midrule
DCI & 71.6 & 61.6 & 51.5 & 31.3 & 38.2 & 45.7 \\
Prompt Optimization & 75.6 & 71.4 & 56.2 & \textbf{41.8} & 38.8 & 52.0 \\
\rowcolor{searchweaverblue}
\Ours{} & \textbf{78.4} & \textbf{74.6} & \textbf{64.6} & 38.3 & \textbf{42.4} & \textbf{55.0} \\
\ \ w/o artifacts and tools & 72.5 & 64.9 & 53.1 & 38.8 & 38.0 & 48.7 \\
\bottomrule
\end{tabularx}
\vspace{-4pt}
\caption{Comparison with prompt optimization and ablation of generated artifacts, tools, and their instructions.}
\label{tab:ablation}
\end{table}

\begin{table}[!t]
\centering
\fontsize{8}{9.5}\selectfont
\setlength{\tabcolsep}{2pt}
\begin{tabularx}{\columnwidth}{@{}l>{\centering\arraybackslash}X>{\hsize=1.1\hsize\centering\arraybackslash}X>{\hsize=0.9\hsize\centering\arraybackslash}X@{}}
\toprule
\multirow{2}{*}[-2.5pt]{Method} & BCP & \multicolumn{2}{c}{BRIGHT} \\
\cmidrule(lr){2-2}\cmidrule(lr){3-4} 
& Acc. (\%) & nDCG@10 & TT (s) \\
\midrule
DCI & 60.9 & 50.7 & 23.1 \\ 
Agentic Search (Dense) & 53.3 & 46.3 & 33.2 \\ 
Agentic Search (BM25) & 44.6 & 38.1 & \textbf{0.4} \\
RISE & 69.1 & 42.4 & 6.6 \\ 
DR-DCI & 69.6 & 47.0 & 4.3 \\ 
\rowcolor{searchweaverblue}
\Ours{} & \textbf{73.9} & \textbf{58.8} & 15.5 \\ 
\bottomrule
\end{tabularx}
\vspace{-4pt}
\caption{Search interface transfer to Gemini~3.5 Flash-Lite without re-optimization.
TT: mean tool execution time per query.}
\label{tab:transfer}
\vspace{-16pt}
\end{table}

\paragraph{Does generating artifacts and tools improve search?}
Table~\ref{tab:ablation} first compares \Ours{} with Prompt Optimization, which uses the same optimization loop as \Ours{} but restricts candidates to instructions, excluding artifacts, tools, and retrieval resources.
Prompt Optimization improves over DCI on both benchmarks, but \Ours{} achieves higher BCP accuracy and average BRIGHT nDCG@10, supporting the value of the broader interface design space.
We also evaluate the optimized interface after removing the generated artifacts and tools and the instructions for using them, without further optimization.
This removal reduces performance on both benchmarks, indicating that the generated artifacts and tools, together with their usage instructions, contribute to the gains.

\paragraph{Do the optimized interfaces transfer across language models?}
We next test whether the discovered interfaces remain useful with another language model.
Table~\ref{tab:transfer} evaluates Gemini~3.5 Flash-Lite~\citep{google2026gemini35flashlite} using the interfaces optimized with GPT-5.4-nano, without re-optimization.
Compared with DCI, the transferred interfaces improve BCP accuracy and average BRIGHT nDCG@10 while reducing mean tool execution time on BRIGHT.
The performance gains hold across all four BRIGHT subsets (Appendix~\ref{sec:transfer-subsets}), suggesting that the optimized interfaces remain useful beyond the model used for optimization.

\clearpage
\section*{Limitations}

Our experiments evaluate interfaces optimized for fixed corpora.
As documents are added or revised, corpus artifacts and the tools that use them may require updates.
We do not study when these updates become necessary or whether they can be performed incrementally without repeating the full optimization process.
Extending \Ours{} to evolving corpora would therefore require mechanisms for maintaining corpus artifacts and assessing whether previously optimized interfaces remain effective.

\bibliography{reference}

\clearpage
\appendix
\section{\Ours{} Algorithm}
\label{sec:optimization-algorithm}

\begin{algorithm}[H]
\caption{\Ours{}: Automatic Search Interface Optimization}
\label{alg:corpus-interface-optimization}
\begin{algorithmic}[1]
\algrenewcommand{\algorithmicrequire}{\textbf{Input:}}
\algrenewcommand{\algorithmicensure}{\textbf{Output:}}
\Require \begin{tabular}[t]{@{}l@{}}
Corpus $C$, queries $\mathcal{Q}$ \\
Fixed agent $f_\theta$, proposer $P$ \\
Initial interface $\phi_0$, iterations $K$
\end{tabular}
\Ensure Optimized interface $\phi^\star$
\Statex $\mathbf{r}$: per-query rewards; $\mathcal{T}$: trajectories
\State $(\mathbf{r}_0, \mathcal{T}_0) \gets \Call{Evaluate}{f_\theta, C, \mathcal{Q}, \phi_0}$
\State $\mathcal{H} \gets \{(\phi_0, \mathbf{r}_0, \mathcal{T}_0)\}$
\State $\phi^\star \gets \phi_0$; $\bar{r}^\star \gets \operatorname{mean}(\mathbf{r}_0)$
\For{$k = 1, \ldots, K$}
  \State $\phi_k \gets \Call{Propose}{P, C, \mathcal{H}}$
  \State Build $\phi_k$'s artifacts and tools over $C$
  \State $(\mathbf{r}_k, \mathcal{T}_k) \gets \Call{Evaluate}{f_\theta, C, \mathcal{Q}, \phi_k}$
  \State $\mathcal{H} \gets \mathcal{H} \cup \{(\phi_k, \mathbf{r}_k, \mathcal{T}_k)\}$
  \If{$\operatorname{mean}(\mathbf{r}_k) > \bar{r}^\star$}
    \State $\phi^\star \gets \phi_k$; $\bar{r}^\star \gets \operatorname{mean}(\mathbf{r}_k)$
  \EndIf
\EndFor
\State \Return $\phi^\star$
\end{algorithmic}
\end{algorithm}

\section{Additional Implementation Details}
\label{sec:experimental-details}

\paragraph{Data and Evaluation.}
On both BCP and BRIGHT, we use the file-system corpus released with DCI-Agent-Lite~\citep{li2026dci}, where each document is stored as a text file.
For BCP, we use 30 queries for interface optimization and 800 for testing.
For BRIGHT, we use 52, 58, 52, and 51 optimization queries and 51, 58, 51, and 50 test queries from Biology, Earth Science, Economics, and Robotics, respectively, totaling 213 optimization and 210 test queries.

\paragraph{Agent Runtime.}
On both BCP and BRIGHT, the search agent runs GPT-5.4-nano with high reasoning effort and has no web access.
Each query is limited to 300 turns and 30 minutes.

\paragraph{Interface Optimization.}
The proposer runs at high reasoning effort and can inspect the corpus, execute shell commands, and call the supplied retrieval resources.
Candidate files must not contain optimization questions, query identifiers, or gold labels; candidates that do are rejected before evaluation.
The context-management policy that a candidate may replace decides how earlier tool outputs are truncated or cleared.
The files containing test queries and labels are made unreadable during optimization.

\paragraph{Retrieval Resources.}
On BCP, \Ours{} uses the official BM25 and Qwen3-Embedding-4B indexes~\citep{chen2025browsecompplus} and encodes dense queries at search time.
On BRIGHT, BM25 uses the official configuration~\citep{su2025bright} and accepts arbitrary queries, whereas the dense resource returns only the released ranking for each original question and does not support new dense queries.

\paragraph{Baseline Configurations.}
\textbf{Agentic Search (Dense)} uses the official BCP~\citep{chen2025browsecompplus} dense-search agent.
On BCP, it also has a tool that returns a full document, and it differs from \Ours{} in both the embedding model and the document index: Agentic Search (Dense) uses Qwen3-Embedding-8B and its corresponding document index, whereas \Ours{} uses Qwen3-Embedding-4B and its corresponding document index.
On BRIGHT, Agentic Search (Dense) has the search tool only and encodes each query at search time against Pyserini's~\citep{lin2021pyserini} prebuilt ReasonEmbed-Qwen3-4B document embeddings.
\textbf{Agentic Search (BM25)} uses the official BCP~\citep{chen2025browsecompplus} function-calling loop, whose only tool returns BM25 snippets; it has no shell or file reader.
On BRIGHT, it uses the same BM25 retriever as \Ours{}.
\textbf{RISE} uses the released harness~\citep{zhuang2026rise} with its released settings.
It adds BM25 results for each subquery to a workspace, restricts shell-based exploration to that workspace, and uses documents augmented in advance by GPT-5.4-nano with tables of contents and section anchors.
\textbf{DR-DCI} uses its released harness~\citep{lu2026drdci} with the configuration of its main BCP result.
On BRIGHT, its pull tool uses the repository's BRIGHT retriever, BGE-large-en-v1.5~\citep{xiao2024cpack}, over Pyserini's prebuilt indexes.

\section{Additional Related Work}

\paragraph{Agentic Search.}
Retrieval-augmented generation grounds language models in external documents through sparse or dense retrieval \citep{lewis2020rag,robertson2009bm25,karpukhin2020dense}.
Agentic approaches make retrieval part of an iterative reasoning process.
Search-R1 learns to interleave reasoning with search-engine calls through reinforcement learning \citep{jin2025searchr1}, while other work explores lexical search, query expansion, and corpus-aware query construction \citep{subramanian2026keyword,hsu2026piserini,yang2026sira}.
Further approaches provide finer control through retrieval actions and logical queries \citep{hui2026interactrag,zeng2026logicalrag}.
DCI and GrepSeek extend this control to direct corpus exploration with terminal commands and raw-file reads, with GrepSeek training agents for this interaction mode \citep{li2026dci,salemi2026grepseek}.
Our work optimizes the interface through which agents combine direct exploration with retrieval tools.

\paragraph{Search Interface Design.}
The tools, document organization, and presentation of search results influence how agents discover and use evidence \citep{sen2026grep,cao2026codingagents}.
RISE constructs bounded interaction spaces, while DR-DCI uses retrieval to dynamically expand a local workspace \citep{zhuang2026rise,lu2026drdci}.
Beyond selecting which documents to expose, other work organizes information into navigable memory hierarchies, source-derived wikis, and environment-side records of requirements and evidence \citep{hsu2026horma,karpathy2026llmwiki,baek2026runningtab}.
We optimize corpus-specific combinations of artifacts, tools, and usage instructions using feedback from search trajectories and task rewards.

\paragraph{Automatic Optimization of Agent Systems.}
Automatic optimization uses task feedback to improve the components surrounding language models, including prompts and multi-stage pipelines \citep{khattab2024dspy,yuksekgonul2025textgrad}.
Recent methods use natural-language feedback and reflection on execution traces to guide revisions, or meta-learning to produce system prompts that generalize across tasks \citep{yuksekgonul2025textgrad,agrawal2026gepa,choi2025metaspo}.
This direction also extends to executable harness code, with coding agents proposing changes based on earlier candidates, scores, and execution traces \citep{lee2026metaharness}.
We apply feedback-driven optimization to search interfaces, jointly developing reusable corpus artifacts, executable search tools, and usage instructions for downstream agentic search.

\section{BRIGHT Tool Time by Subset}
\label{sec:bright-cost}

Table~\ref{tab:bright-cost} reports tool time, the total execution time of a query's tool calls, excluding model inference.
Aggregate BRIGHT timings are computed over all 210 test queries.

\begin{table}[!t]
\centering
\fontsize{8}{9.5}\selectfont
\setlength{\tabcolsep}{1.5pt}
\begin{tabularx}{\columnwidth}{@{}ll*{3}{>{\raggedleft\arraybackslash}X}@{}}
\toprule 
& & \multicolumn{3}{c}{Tool time (s)} \\
\cmidrule(lr){3-5}
Subset & Method & Mean & Median & P95 \\
\midrule
\multirow{6}{*}{Biology} & DCI & 62.2 & 37.5 & 250.7 \\ 
& Agentic Search (Dense) & 40.9 & 42.3 & 75.4 \\ 
& Agentic Search (BM25) & 0.3 & 0.2 & 0.6 \\ 
& RISE & 1.7 & 1.5 & 5.2 \\ 
& DR-DCI & 8.0 & 5.5 & 20.8 \\ 
& \cellcolor{searchweaverblue}\Ours{} & \cellcolor{searchweaverblue}12.3 & \cellcolor{searchweaverblue}9.7 & \cellcolor{searchweaverblue}28.5 \\
\midrule
\multirow{6}{*}{\shortstack[l]{Earth\\Science}} & DCI & 78.8 & 70.1 & 166.7 \\ 
& Agentic Search (Dense) & 40.1 & 35.2 & 109.7 \\ 
& Agentic Search (BM25) & 0.5 & 0.4 & 1.1 \\ 
& RISE & 2.2 & 1.3 & 8.6 \\ 
& DR-DCI & 9.8 & 6.0 & 27.3 \\ 
& \cellcolor{searchweaverblue}\Ours{} & \cellcolor{searchweaverblue}63.5 & \cellcolor{searchweaverblue}51.4 & \cellcolor{searchweaverblue}162.5 \\
\midrule
\multirow{6}{*}{Economics} & DCI & 43.3 & 28.4 & 144.0 \\ 
& Agentic Search (Dense) & 41.7 & 39.9 & 79.9 \\ 
& Agentic Search (BM25) & 0.3 & 0.3 & 0.6 \\ 
& RISE & 1.9 & 1.4 & 6.5 \\ 
& DR-DCI & 10.1 & 7.5 & 25.5 \\ 
& \cellcolor{searchweaverblue}\Ours{} & \cellcolor{searchweaverblue}26.3 & \cellcolor{searchweaverblue}20.3 & \cellcolor{searchweaverblue}73.3 \\
\midrule
\multirow{6}{*}{Robotics} & DCI & 66.6 & 46.8 & 214.8 \\ 
& Agentic Search (Dense) & 45.0 & 43.0 & 98.8 \\ 
& Agentic Search (BM25) & 0.3 & 0.3 & 0.6 \\ 
& RISE & 1.7 & 1.3 & 4.7 \\ 
& DR-DCI & 17.2 & 8.7 & 65.8 \\ 
& \cellcolor{searchweaverblue}\Ours{} & \cellcolor{searchweaverblue}40.7 & \cellcolor{searchweaverblue}38.8 & \cellcolor{searchweaverblue}100.7 \\
\midrule
\multirow{6}{*}{All} & DCI & 63.2 & 47.1 & 166.7 \\ 
& Agentic Search (Dense) & 41.8 & 38.7 & 82.9 \\ 
& Agentic Search (BM25) & 0.4 & 0.3 & 0.8 \\ 
& RISE & 1.9 & 1.4 & 6.1 \\ 
& DR-DCI & 11.2 & 7.1 & 31.8 \\ 
& \cellcolor{searchweaverblue}\Ours{} & \cellcolor{searchweaverblue}36.6 & \cellcolor{searchweaverblue}25.8 & \cellcolor{searchweaverblue}103.8 \\
\bottomrule
\end{tabularx}
\caption{BRIGHT tool time per test query by subset (GPT-5.4-nano). P95 is the nearest-rank 95th percentile. Agentic Search (Dense) excludes the time a search waits for other queries' searches on the shared CPU encoder.}
\label{tab:bright-cost}
\end{table}

\section{BRIGHT Transfer Results by Subset}
\label{sec:transfer-subsets}

Table~\ref{tab:transfer-subsets} reports BRIGHT nDCG@10 by subset for the transfer to Gemini~3.5 Flash-Lite (Table~\ref{tab:transfer}).

\begin{table}[!t]
\centering
\fontsize{8}{9.5}\selectfont
\setlength{\tabcolsep}{1.5pt}
\begin{tabularx}{\columnwidth}{@{}l*{5}{>{\centering\arraybackslash}X}@{}}
\toprule
Method & Bio. & Earth & Econ. & Rob. & Avg. \\
\midrule
DCI & 63.1 & 55.7 & 38.5 & 45.6 & 50.7 \\
Agentic Search (Dense) & 57.6 & 56.6 & 36.4 & 34.4 & 46.3 \\
Agentic Search (BM25) & 47.8 & 40.1 & 31.1 & 33.4 & 38.1 \\
RISE & 57.6 & 49.8 & 27.5 & 34.8 & 42.4 \\
DR-DCI & 60.4 & 56.0 & 34.9 & 36.5 & 47.0 \\
\rowcolor{searchweaverblue}
\Ours{} & \textbf{75.2} & \textbf{66.0} & \textbf{45.1} & \textbf{48.6} & \textbf{58.8} \\
\bottomrule
\end{tabularx}
\caption{BRIGHT nDCG@10 by subset with Gemini~3.5 Flash-Lite, using the interfaces optimized with GPT-5.4-nano.}
\label{tab:transfer-subsets}
\end{table}

\section{BCP Agent Trajectory}
\label{sec:bcp-trajectory}

\begin{figure*}[p]
  \centering
  \includegraphics[width=\textwidth,height=.95\textheight,keepaspectratio]{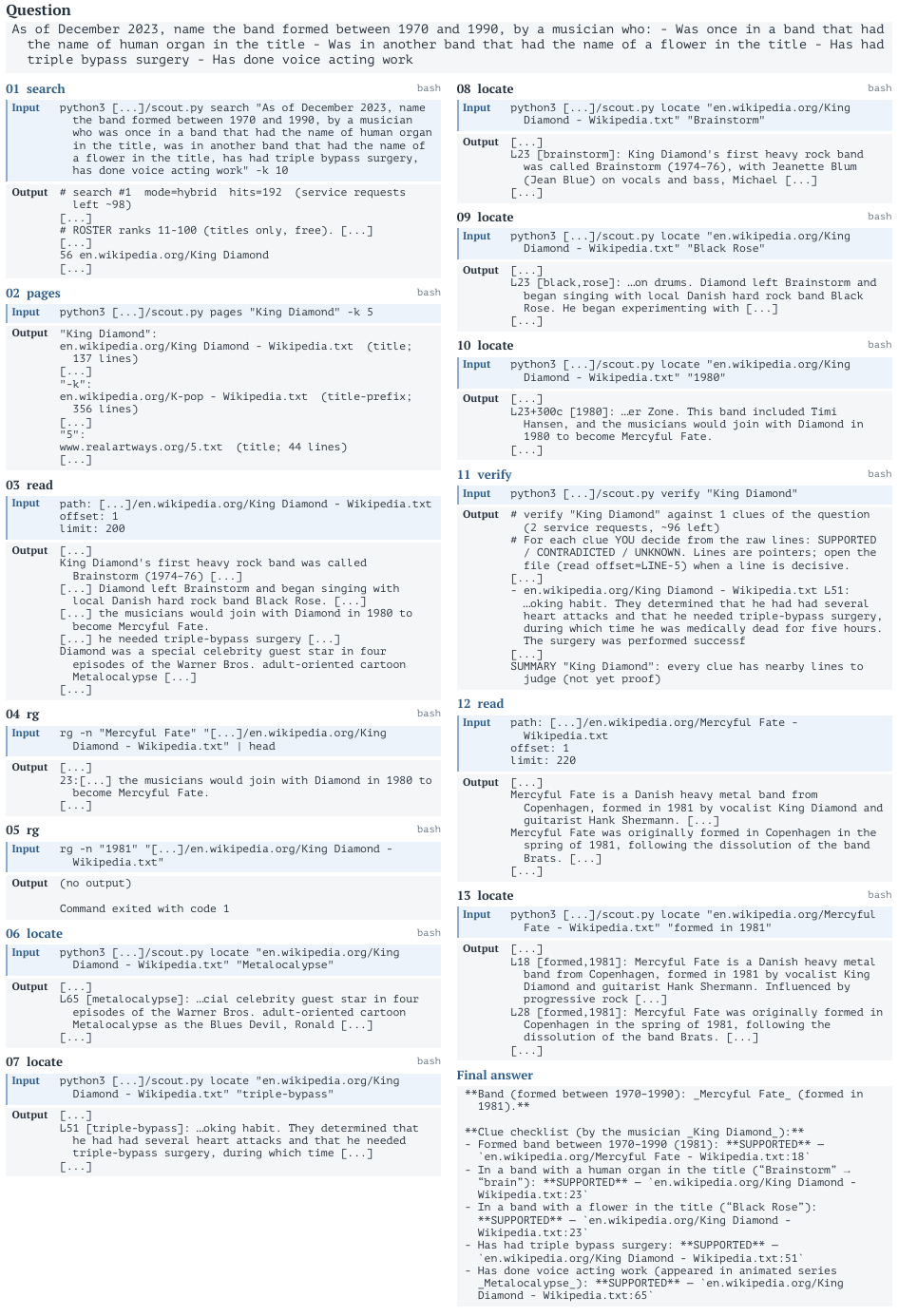}
  \caption{Example of a full agent trajectory using the optimized search interface on BrowseComp-Plus.
  All 13 tool calls and the final answer are shown; outputs are abbreviated.}
  \label{fig:bcp-full-trajectory}
\end{figure*}

Figure~\ref{fig:bcp-full-trajectory} presents all 13 tool calls and the final answer for the test query in Figure~\ref{fig:bcp-qualitative}.
Outputs and paths are abbreviated for readability.

\section{Optimized Interfaces on BRIGHT}
\label{sec:bright-interfaces}

\begin{figure*}[p]
\centering
\begin{minipage}[t]{0.43\textwidth}\vspace{0pt}\centering
\includegraphics[width=\linewidth]{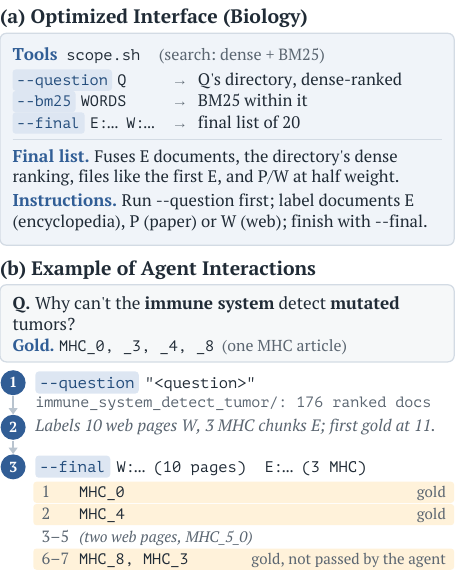}
\end{minipage}\hfill
\begin{minipage}[t]{0.43\textwidth}\vspace{0pt}\centering
\includegraphics[width=\linewidth]{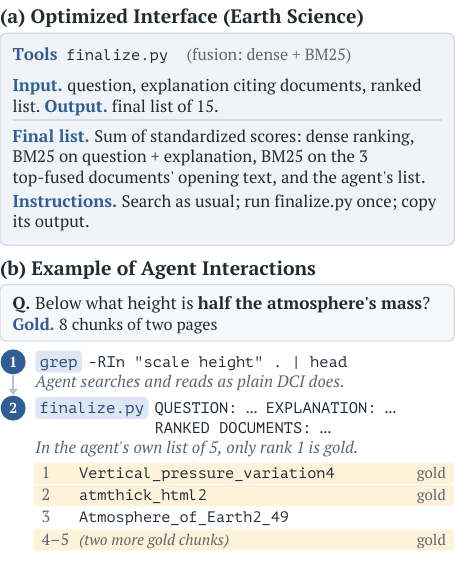}
\end{minipage}

\begin{minipage}[t]{0.43\textwidth}
\caption{\Ours{}'s optimized interface for BRIGHT Biology. The example in (b) scores 90.6 nDCG@10.}
\label{fig:interface-biology}
\end{minipage}\hfill
\begin{minipage}[t]{0.43\textwidth}
\caption{\Ours{}'s optimized interface for BRIGHT Earth Science. The example in (b) scores 61.9 nDCG@10.}
\label{fig:interface-earth-science}
\end{minipage}

\vspace{1.2em}
\begin{minipage}[t]{0.43\textwidth}\vspace{0pt}\centering
\includegraphics[width=\linewidth]{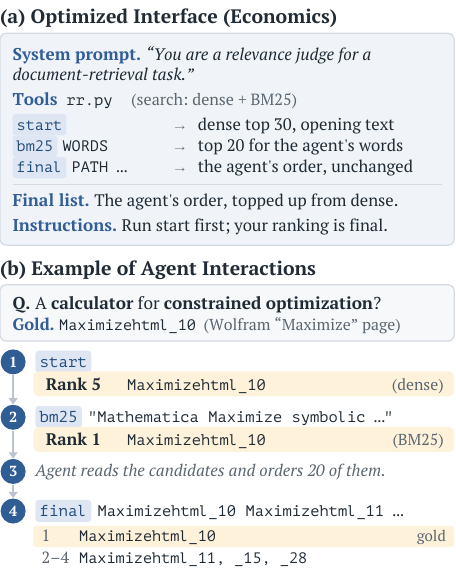}
\end{minipage}\hfill
\begin{minipage}[t]{0.43\textwidth}\vspace{0pt}\centering
\includegraphics[width=\linewidth]{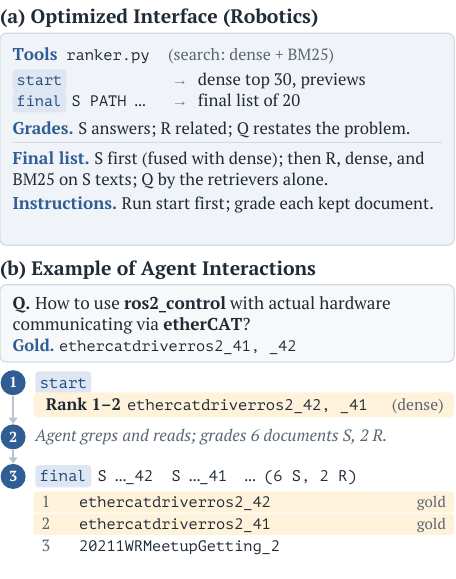}
\end{minipage}

\begin{minipage}[t]{0.43\textwidth}
\caption{\Ours{}'s optimized interface for BRIGHT Economics. The example in (b) scores 100.0 nDCG@10.}
\label{fig:interface-economics}
\end{minipage}\hfill
\begin{minipage}[t]{0.43\textwidth}
\caption{\Ours{}'s optimized interface for BRIGHT Robotics. The example in (b) scores 100.0 nDCG@10.}
\label{fig:interface-robotics}
\end{minipage}
\end{figure*}

Figures~\ref{fig:interface-biology}--\ref{fig:interface-robotics} show the interface \Ours{} selected for each BRIGHT subset: panel (a) summarizes its tools, how its final tool builds the submitted list, and its instructions, and panel (b) traces one test question from the agent's tool calls to the submitted list, with gold documents highlighted.

The four interfaces share one structure.
The agent still searches and reads raw files with the shell, and a final tool builds the submitted list by combining the agent's judgments with the released dense ranking; Earth Science and Robotics also mix into that list BM25 scores computed from text the agent has read or written.
The interfaces differ in where retrieval enters and how much weight the agent's judgments receive.
Biology confines the search to the question's topic directory and fuses the documents the agent labels with the dense ranking inside that directory.
Earth Science searches as plain DCI does and uses retrieval only in the final tool.
Economics and Robotics start from dense candidates: Economics keeps the agent's order and fills the remaining places from the dense ranking, while Robotics ranks the documents the agent grades as answering the question first.

Removing the generated artifacts and tools hurts most on Biology and Earth Science, where \Ours{} gains most.
Biology and Earth Science improve over DCI by 13.0 and 13.1 points and lose 9.7 and 11.5 points without the artifacts and tools (Table~\ref{tab:ablation}).
Economics improves by 7.0 points and does not drop without them (38.8 vs.\ 38.3), and Prompt Optimization scores highest there (41.8).

\clearpage

\end{document}